\documentclass[final]{cvpr}

\usepackage{times}
\usepackage{epsfig}
\usepackage{graphicx}
\usepackage{amsmath}
\usepackage{amssymb}

\usepackage{array}
\usepackage{setspace}
\usepackage{multirow}
\usepackage{makecell}
\usepackage{subfloat}
\usepackage{booktabs} 

\usepackage{multirow}

\usepackage[pagebackref=true,breaklinks=true,colorlinks,bookmarks=false]{hyperref}
\def\cvprPaperID{4139} % *** Enter the CVPR Paper ID here

\def\cvprPaperID{4139} % *** Enter the CVPR Paper ID here
\def\confYear{CVPR 2021}

\begin{document}

\title{VoxelContext-Net: An Octree based Framework for Point Cloud Compression}  % **** Enter the paper title here

\author{Zizheng Que\\
Beihang University\\
{\tt\small quezizheng@buaa.edu.cn}
\and
Guo Lu\thanks{Guo Lu is the corresponding author.}\\
Beijing Institute of Technology\\
{\tt\small guo.lu@bit.edu.cn}
\and
Dong Xu\\
University of Sydney\\
{\tt\small dong.xu@sydney.edu.au}
}

\maketitle
\thispagestyle{empty}

%%%%%%%%% BODY TEXT - ENTER YOUR RESPONSE BELOW
\begin{abstract}

In this paper, we propose a two-stage deep learning framework called VoxelContext-Net for both static and dynamic point cloud compression. 
Taking advantages of both octree based methods and voxel based schemes, our approach employs the voxel context to compress the octree structured data. 
Specifically, we first extract the local voxel representation that encodes the spatial neighbouring context information for each node in the constructed octree. 
Then, in the entropy coding stage, we propose a voxel context based deep entropy model to compress the symbols of non-leaf nodes in a lossless way. 
Furthermore, for dynamic point cloud compression, we additionally introduce the local voxel representations from the temporal neighbouring point clouds to exploit temporal dependency. 
More importantly, to alleviate the distortion from the octree construction procedure, we propose a voxel context based 3D coordinate refinement method to produce more accurate reconstructed point cloud at the decoder side, which is applicable to both static and dynamic point cloud compression.
The comprehensive experiments on both static and dynamic point cloud benchmark datasets(\textit{e.g.,} ScanNet and Semantic KITTI) clearly demonstrate the effectiveness of our newly proposed method VoxelContext-Net for 3D point cloud geometry compression.

\end{abstract}

\section{Introduction}
\label{sec:introduction}

Due to the rapid population of 3D sensors such as LiDAR, there is an increasing research interest to compress tremendous amount of 3D point cloud data for a broad range of applications~(\textit{e.g.,} autonomous driving).
When compared with image and video compression~\cite{wallace1992jpeg,taubman2002jpeg2000,bellard2015bpg,wiegand2003overview,sullivan2012overview,lu2020content}, it is a more challenging task to compress a set of orderless 3D points from point clouds.

\begin{figure}[t]
\centering
\begin{minipage}[c]{\linewidth}
\includegraphics[width=\linewidth]{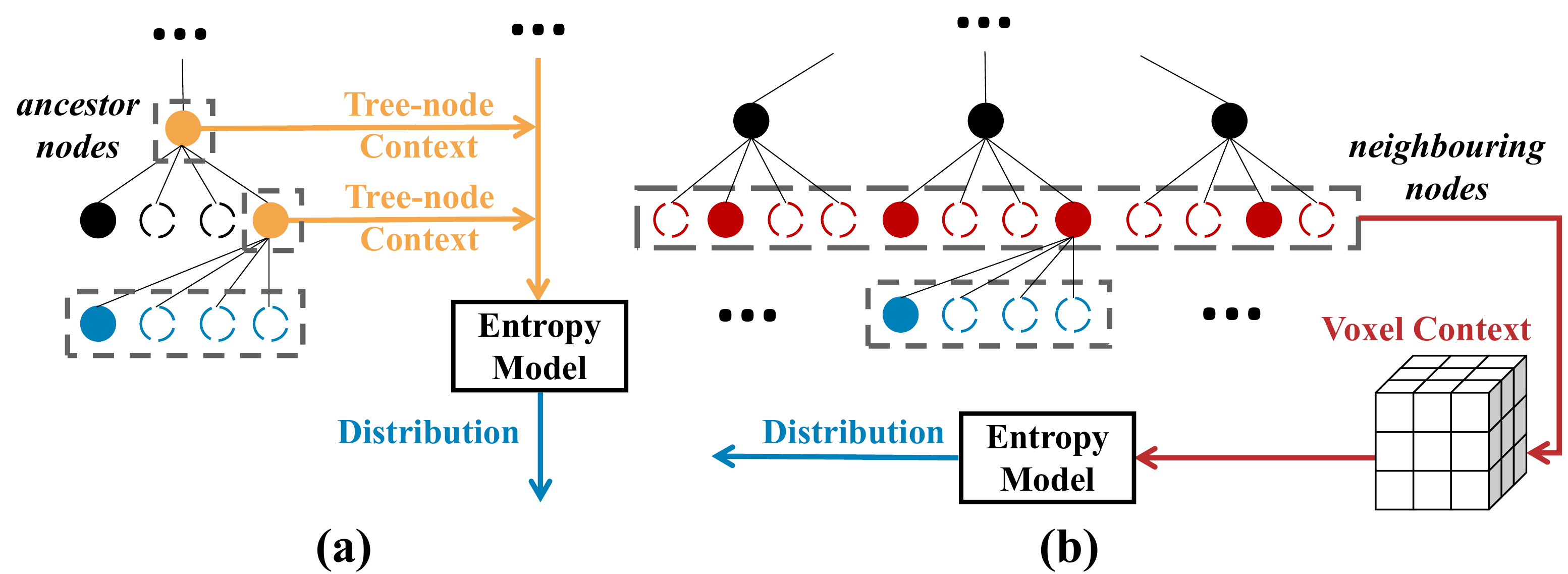}
\end{minipage}
\caption{Comparison between two context generation approaches for learning the entropy model based on the octree structured data. (a) Tree-node context used in~\cite{Huang_2020_CVPR}.  (b) Local voxel context employed in our approach.}
\label{fig:Different_Context}
\end{figure}

Recently, several deep learning methods were developed for point cloud compression. For example, Wang \textit{et al.}~\cite{wang2019learned} transformed the point cloud data to the voxel representation in order to capture the spatial dependency, and then they employed the existing image compression method~\cite{balle2018variational} for point cloud compression. Other recent works \cite{yan2019deep} and \cite{huang20193d} directly compressed the raw point cloud data by using the existing backbone networks~(\textit{e.g.,} PointNet/PointNet++~\cite{qi2017pointnet,qi2017pointnet++}) for feature encoding.
These voxel-based or point-based methods can take advantage of the existing image compression or point cloud analysis techniques. However, the voxel-based methods ignore the sparsity characteristic of point clouds and thus suffer from the relatively high computational complexity~\cite{wang2019learned} while the point based methods are inefficient when processing large point cloud data~\cite{yan2019deep}.
In~\cite{Huang_2020_CVPR}, Huang \textit{et al.} used the octree to organize the point cloud data and proposed an entropy model to exploit the dependency between multiple parent nodes and each child node~(see Figure~\ref{fig:Different_Context}(a)). Although their approach inherits the benefits for efficiently processing octree structured point cloud data, the strong dependency among the neighboring octree nodes at the same depth level is unfortunately ignored in their octree-based entropy model~\cite{Huang_2020_CVPR}. Furthermore, the distortions are also introduced after converting the raw point cloud to the octree structured data, which further degrades the compression performance.
Besides, their approach is only designed for static point cloud compression, which may limit the performance for dynamic point cloud compression.

To address these issues, we propose a new learning based point cloud compression method VoxelContext-Net by exploiting the voxel context in the octree based framework.
Our approach takes advantage of the efficient data organization ability from the octree based methods and the spatial modeling capability from the voxel based methods,  which can be applied to both static and dynamic point cloud geometry compression.
Specifically, the input point cloud is first organized by using the octree structure, where the symbol of each non-leaf node represents the occupancy status of its eight children. 
In the entropy coding stage, we propose a new learning based entropy model to compress these symbols in a lossless way.
To effectively produce context information for the entropy model, we exploit the local binary voxel representation for each node, where the entries of our voxel representation indicate the existence of neighbouring nodes at the same depth level~(see Figure~\ref{fig:Different_Context}(b)).
Furthermore, to reduce temporal redundancy for dynamic point cloud compression, we additionally include the co-located voxel representations from the previous and the subsequent point clouds to generate richer context information. 
In the reconstruction stage,  we further propose a coordinate refinement method based on the local voxel representations at the decoder side to produce more accurate 3D coordinate for each leaf node in both static and dynamic point clouds.

We evaluate the performance of our newly proposed method on the large-scale 3D static and dynamic point cloud datasets~(\textit{e.g.,} ScanNet~\cite{dai2017scannet} and Semantic KITTI~\cite{geiger2012cvpr,behley2019iccv}). The comprehensive experiments demonstrate that our method outperforms both hand-crafted point cloud compression methods and the learning-based point cloud compression methods.

The contributions of our work are highlighted as follows: 

\begin{itemize}
\item
 By taking the advantage of both voxel based methods and octree based schemes, we introduce local voxel context in the deep entropy model for better compression of octree structured data. Our approach can be applied to both static and dynamic point cloud compression.

\item We develop a voxel context based coordinate refinement module to produce accurate coordinates of leaf nodes at the decoder side.

\item Our simple and effective approach achieves the state-of-the-art compression performance on several large-scale datasets for both static and dynamic point cloud geometry compression. 

\end{itemize}

\section{Related Work}

\subsection{Traditional Point Cloud Compression Methods}

In the past several years, a few point cloud compression methods~\cite{Rusu_ICRA2011_PCL,google,schwarz2018emerging,jackins1980oct,schnabel2006octree,de2016compression,de2017transform} have been proposed and most of them are based on the tree representations. 
For example, the MPEG group developed a standard point cloud compression method~\cite{schwarz2018emerging,gpcc,gpcc} G-PCC~(geometry based point cloud compression) for static point clouds, which includes an octree-structure based method for point cloud compression. 
However, they are all based on hand-crafted techniques and thus cannot be optimized in an end-to-end fashion by using large-scale data.

In addition, although some learning based image and video compression approaches~\cite{balle2016end,balle2018variational,minnen2018joint,lu2019dvc,Wu_2018_ECCV,lu2020end,hu2020improving} have been proposed, it is still a non-trial task to employ the standard CNN operations for compressing point clouds consisting of a sparse set of orderless 3D points.

\subsection{Deep Learning for Point Cloud Compression}

Taking the point cloud data as the input, Yan \textit{et al.}~\cite{yan2019deep} built an auto-encoder network by using PointNet, in which the latent representation is quantized and further compressed by using an entropy coding model.
These point based methods~\cite{yan2019deep,huang20193d} may suffer from the huge memory usage issue and high computational costs.
Wang \textit{et al.}~\cite{wang2019learned} extended the existing image compression method~\cite{balle2016end} for voxelized point cloud compression. 
Unfortunately, their approach ignores the sparsity characteristic of point clouds and thus the computational complexity is relatively high when compared with the octree based methods. 

Recently, an octree-based method OctSqueeze~\cite{Huang_2020_CVPR} was proposed. While the OctSqueeze method~\cite{Huang_2020_CVPR} avoids the issues related to high memory usage and slow encoding/decoding speed, their approach still suffer from the following drawbacks.
First, they only exploited context information from its \textit{ancestor nodes}~(as shown in Figure~\ref{fig:Different_Context}(a)) to predict the probability model in the entropy model, which ignores strong prior information between \textit{spatial neighbouring nodes} at the same depth level. 
In addition, their work does not consider the distortion introduced in the octree construction procedure, and their method is only designed for static point cloud compression.
Although there is a concurrent work~\cite{biswas2020muscle} for dynamic point cloud compression, it follows the existing framework from \cite{Huang_2020_CVPR} and thus suffers from similar limitations.

In contrast to these works~\cite{yan2019deep,huang20193d,wang2019learned,Huang_2020_CVPR,biswas2020muscle,tu2019real}, we propose to exploit context information between neighbouring nodes by using local voxel representation in our deep entropy model and our work also refines the 3D coordinate at the decoder side in order to achieve better reconstruction results. 
Besides, we further extend our approach by additionally exploit the local context representation from neighbour frames for dynamic point cloud compression.

\begin{figure}[t]
\centering
\includegraphics[width=\linewidth]{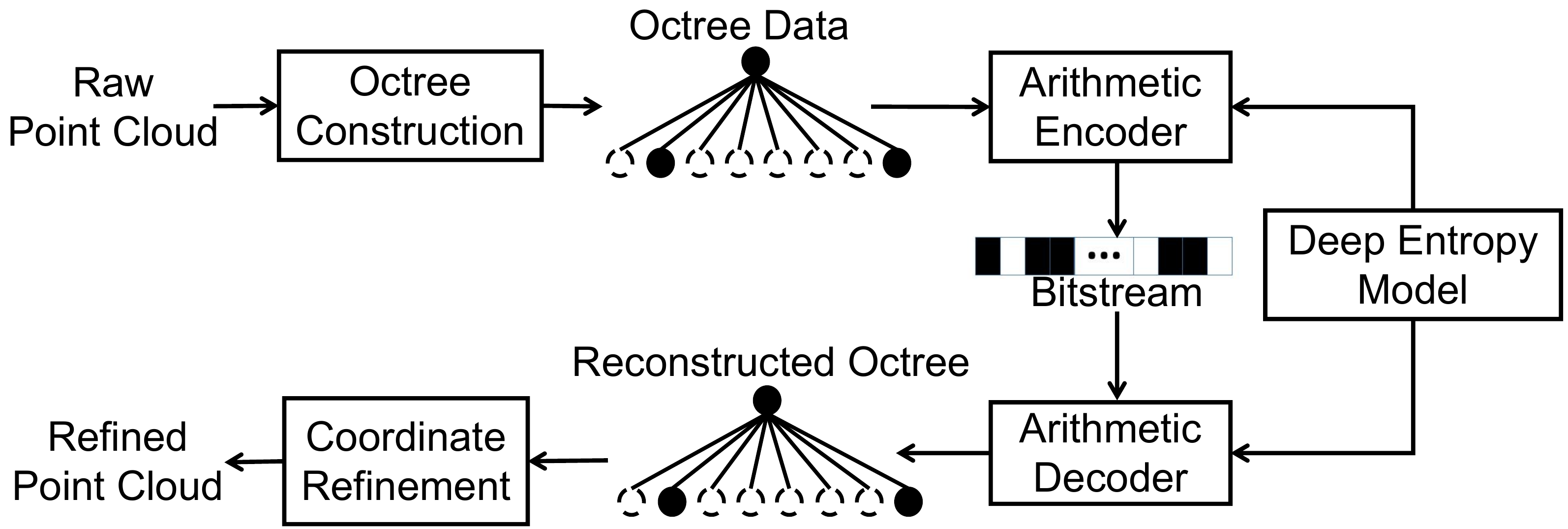}
\caption{The overall architecture of our proposed static point cloud compression approach. The symbols of the non-leaf nodes from the octree are losslessly compressed by using the proposed deep entropy model, while the coordinate refinement module is used to predict more accurate coordinates at the decoder side.}
\label{fig:WholeNetwork}
\end{figure}

\section{Methodology}

\subsection{Overview}
The overall architecture of the proposed point cloud compression method is shown in Fig~\ref{fig:WholeNetwork}. 
In this section, we first take the static point cloud compression as an example to illustrate our proposed method and then introduce how to extend the proposed method for dynamic point cloud compression.

Specifically, in the first stage, we organize the input static point cloud by using the octree structure, in which our approach aims to encode these symbols of non-leaf nodes in a lossless way. 
To improve the compression performance, we propose a voxel context based deep entropy model to accurately predict the probability distributions of these symbols.
Furthermore, to compensate the distortion in the octree construction procedure, a local voxel context based coordinate refinement module is proposed to produce more accurate reconstructed point cloud at the decoder side.

\subsection{Octree Construction}

In Figure~\ref{fig:voxel}, we provide a toy example to illustrate the octree construction procedure.
% , where Figure~\ref{fig:voxel}(a) depicts the sparse 3D point cloud and Figure~\ref{fig:voxel}(b) represents the corresponding octree. 
Specifically, an octree can be constructed from any 3D point cloud by first partitioning the 3D space into 8 cubes with the same size, and then recursively partitioning each non-empty cube in the same way until the maximum depth level is reached. 
The 3D coordinate of each node represents the cube center.
For each non-leaf node, a 8-bit symbol is used to represent the occupancy status of its eight children, with each bit corresponding to one specific child.

In the octree construction procedure, the quality of the reconstructed point cloud depends on the maximum depth level in the octree structure. 
Therefore, the coordinate of the current leaf node~(\textit{i.e.,} the cube center) is not always consistent with the original 3D coordinate of the corresponding point in the raw point cloud. 
For example, the coordinate of one input point $r_i$ is $(0.6,0.7,0.7)$, while the coordinate for the corresponding leaf node $n_i$ is quantized to $(0.625,0.625,0.625)$, thus inevitable distortion is introduced in the octree construction procedure.  
In this work, we will compress the symbols of octree nodes losslessly and recover the accurate decoded coordinates at the decoder side.

\subsection{Local Voxel Context in Octree}
\label{sec:conetxt}

In the octree structure, the parent node will generate 8 children nodes, which is equivalent to bisecting the 3D space along x-axis, y-axis and z-axis. 
Thus, the partition of the original space at the $k$th depth level in the octree is equivalent to dividing the corresponding 3D space $2^k$ times along x-axis, y-axis and z-axis, respectively. 
Then we produce a binary voxel representation with the shape of $2^{k} * 2^{k} * 2^{k}$  based on the existence of points in each cube. 
Here we assume the corresponding local voxel representations centered at the node $n_i$ is $\boldsymbol{V_i} \in \mathcal{R}^{M\times M \times M}$, where $M$ represents the size of the local voxel representation.
$\boldsymbol{V_i}$ will be used in our approach as strong prior information to improve the compression performance.

\begin{figure}[t]
\centering
\includegraphics[width=0.8\linewidth]{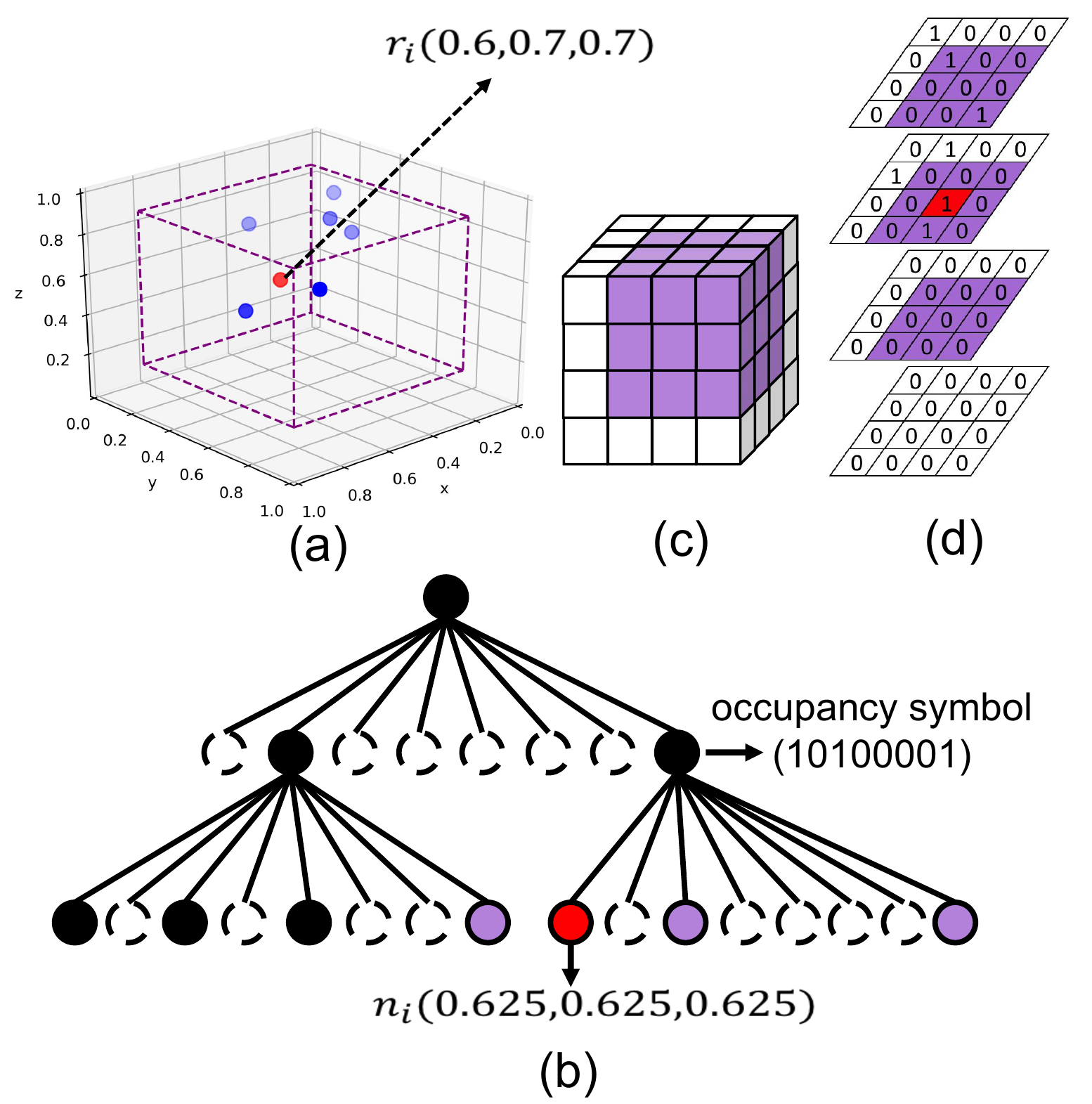}
\caption{A toy example for constructing the octree and extracting the local voxel context representation. (a) Raw input point cloud. (b) The corresponding octree. (c) The voxel representation for the input point cloud at the depth level of 2. (d) The detailed binary voxel representation. 
}
\label{fig:voxel}
\end{figure}

In Figure~\ref{fig:voxel}(c), the purple region represents the local voxel context for the current node $n_i$ and the detailed binary values for the local voxel representation are depicted in Figure~\ref{fig:voxel}(d). In Figure~\ref{fig:voxel}(a), we also provide the 3D coordinate range of the corresponding local region in the 3D space(see the purple dash line).
It is noted that the local voxel context $\boldsymbol{V_i}$ of the current node $n_i$ represents the distribution information of neighbouring nodes at the same depth level.
In contrast, the previous method~\cite{Huang_2020_CVPR} only exploit the information from its ancestor nodes without considering strong spatial neighbouring prior information(see Figure~\ref{fig:Different_Context}(a)). 

\subsection{Our Deep Entropy Model}

\subsubsection{Formulation}
Let $\boldsymbol{s} = [\boldsymbol{s_1}, . . .,\boldsymbol{s_i}, . . .]$ denote a sequence of 8-bit occupancy symbols from all non-leaf octree nodes where $\boldsymbol{s_i}$ represents the symbol for node $n_i$ in the octree.
For example, when $\boldsymbol{s_i} = [0,0,0,1,0,0,0,1]$, it means node $n_i$ has two children and the corresponding indexes of the two children are 4 and 8, respectively.

According to the information theory~\cite{shannon1948mathematical}, when compressing the occupancy information, the lower bound of bitrates is the Shannon entropy.
However, the actual distribution $P$ is unknown in the practical applications.
Therefore, we use the deep neural network to estimate the probability distribution which can be employed to approximate the actual distribution $P$.
Based on the learned deep entropy model, we can compress these occupancy symbols from the octree in a lossless manner. 
Specifically, the objective of our learning based entropy model is to minimize the  cross-entropy loss $E_{\boldsymbol{s} \sim P}[-{\rm log} Q_s(\boldsymbol{s})]$, where $Q_s(\boldsymbol{s})$ is the estimated probability of $\boldsymbol{s}$.

It is noted that the children's probability distribution $q_s(\boldsymbol{s_i})$ at the current node $n_i$ 
may rely on the previously decoded nodes as well as the neighbouring nodes at the current depth level, so it is very difficult to model this complex relationship.
In this work, we assume the occupancy symbol $\boldsymbol{s_i}$ for the current node $n_i$ only depends on the node's local voxel context $\boldsymbol{V_i}$ and the node feature $\boldsymbol{c_i}$ that includes the node's 3D coordinate and the depth level of the node. Since $\boldsymbol{V_i}$ represents local context information at the same depth level, it is reasonable to infer the node $n_i$'s children distribution~(\textit{i.e.,} $\boldsymbol{s_i}$) based on $\boldsymbol{V_i}$.
Then we can simplify the complex dependence relationship for each original occupancy symbol $\boldsymbol{s}$ and further factorize $Q_s(\boldsymbol{s})$ into the following way,
\begin{equation}
\setlength{\abovedisplayskip}{0.15cm}
\setlength{\belowdisplayskip}{0.1cm}
Q_s(\boldsymbol{s}) = \prod_{i}q_s(\boldsymbol{s_i} | \boldsymbol{V_i, c_i})
\label{eq:2}
\end{equation}
where $q_s(\boldsymbol{s_i}|\boldsymbol{V_i},\boldsymbol{c_i})$ is the predicted probability of $\boldsymbol{s_i}$ when the local voxel context $\boldsymbol{V_i}$ and the node feature $\boldsymbol{c_i}$ are given at node $n_i$.
Here, we use the same probability distribution model for each node.

\begin{figure}[t]
\centering
\includegraphics[width=\linewidth]{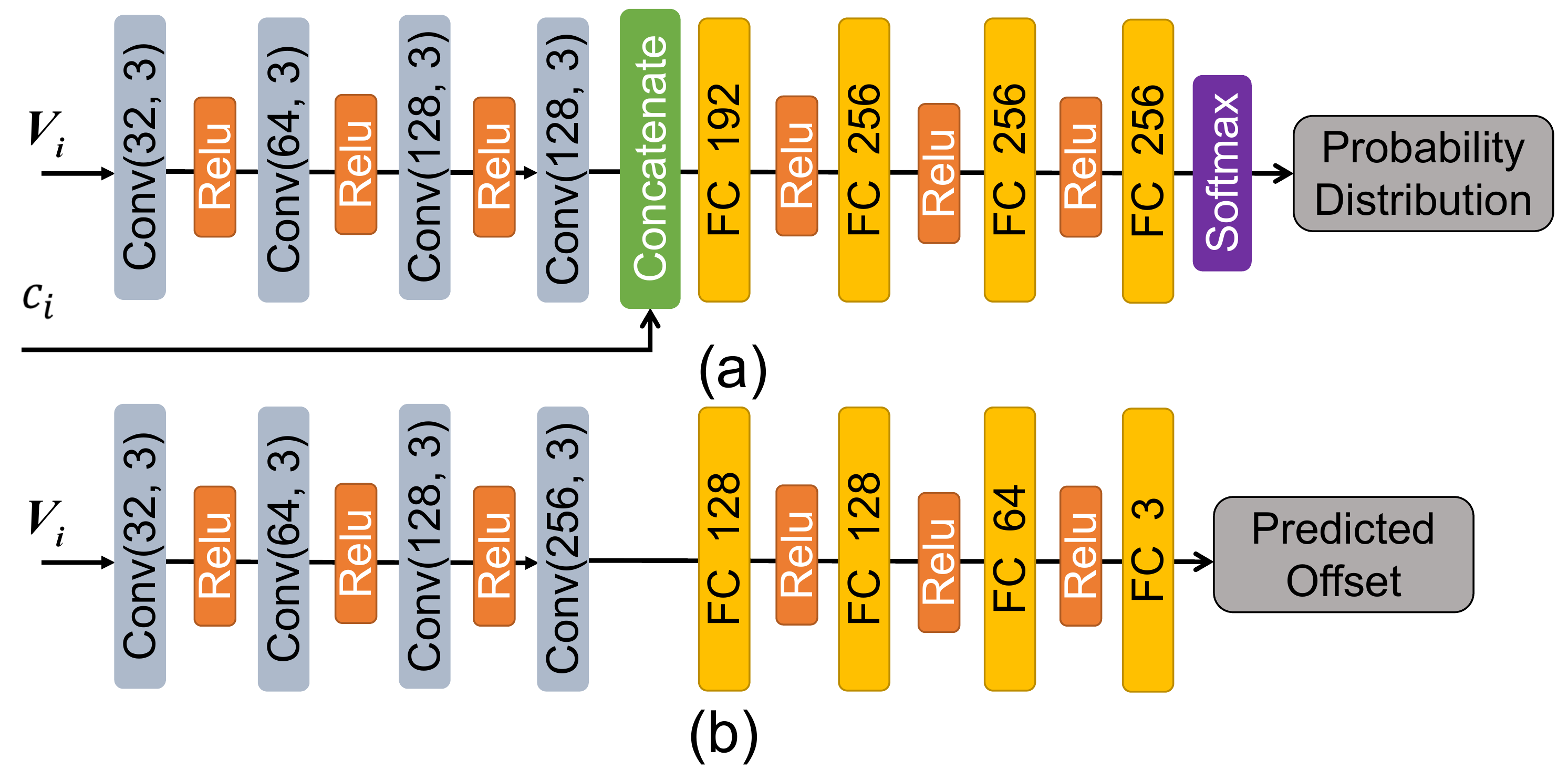}
\caption{Network architecture of (a) our proposed deep entropy model and (b) our coordinate refinement module. `Conv(128,3)' represents the 3D convolution operation when the number of channels is 128 and the kernel size is $3 \times 3 \times 3$.}
\label{fig:dequantization}
\end{figure}

\vspace{-2mm}

\subsubsection{Network Architecture}
In this work, we use deep neural networks to parameterize the entropy model in Eq.~(\ref{eq:2}). 
The whole network architecture is shown in Figure~\ref{fig:dequantization}.
Specifically, for the current node $n_i$, we first extract its local voxel context $\boldsymbol{V_i}$ based on the method described in Section~\ref{sec:conetxt}.
Then $\boldsymbol{V_i}$ is fed to a multi-layer convolutional neural networks (CNN) and the corresponding output is $\boldsymbol{f_i}$. 
In this procedure, we adopt the popular and off-shelf CNN structure to effectively exploit context information in the 3D space, which has not been exploited by the existing octree based point cloud compression methods.
Then we concatenate the context feature $\boldsymbol{f_i}$ and the node feature $\boldsymbol{c_i}$.
After that, a multi-layers perception (MLPs) is adopted to generate a 256-dimensional hidden feature, which fuses both local voxel contextual information and node information. 
Finally, a softmax layer is used to produce the probabilities $q_s(\boldsymbol{s_i}|\boldsymbol{V_i},\boldsymbol{c_i})$ of the 8-bit occupancy symbol for each given node $n_i$.
Our approach takes advantage of both octree based methods and voxel based methods and achieves better point cloud compression performance.

\subsection{Our Coordinate Refinement Method}
\label{sec:CR}

To reduce the distortion in the octree construction procedure, we propose a local voxel context based coordinate refinement method to produce more accurate reconstructed point cloud at the decoder side.
A key intuition behind our coordinate refinement method is that we can better predict the coordinate of the current node by exploiting local context information of its neighbouring voxels.  
 
 Given the decoded octree, for each leaf node $n_i$, the goal of our coordinate refinement method is to predict the refined output coordinate $(x_i^{r}, y_i^{r}, z_i^{r})$ in the following way,
\begin{equation}
\setlength{\abovedisplayskip}{0.15cm}
\setlength{\belowdisplayskip}{0.1cm}
\label{eq:3}
 (x_i^{r}, y_i^{r}, z_i^{r}) = (x_i^d, y_i^d, z_i^d) +  R(\boldsymbol{V_i})
\end{equation}
where $\boldsymbol{V_i}$ is the local voxel context for the leaf node $n_i$ and $R(\cdot)$ is a learnable function for coordinate refinement. The decoded coordinate $(x_i^{d}, y_i^{d}, z_i^{d})$ represents the coordinate of node $n_i$ after octree decoding. 

As shown in Figure~\ref{fig:dequantization}(b), the network architecture of the proposed coordinate refinement method is similar to the deep entropy model.
Specifically, for a decoded octree, we firstly transform the whole octree into a global binary voxel representation, from which we can readily produce the local voxel context for each leaf node. Given the binary voxel context $\boldsymbol{V_i}$ for node $n_i$, we firstly extract context information in the 3D space by using multi-layer CNNs, and then predict the offset by using a set of fully connected layers.
We then calculate the decoded coordinate of the leaf node $n_i$ and the final reconstructed coordinate of $n_i$ is the sum of its decoded coordinate and the predicted offset~(see Eq.~(\ref{eq:3})).

\subsection{Training Strategy}
In the proposed method, we separately train our local voxel context based deep entropy model and our coordinate refinement module. 
For the deep entropy model,  based on the predicted distribution $q_s(\boldsymbol{s_i}|\boldsymbol{V_i},\boldsymbol{c_i})$ at each non-leaf node, we use the following loss function:
\begin{equation}
\setlength{\abovedisplayskip}{0.15cm}
\setlength{\belowdisplayskip}{0.1cm}
L_{entropy} = -\sum_{i} {\rm log}q_s(\boldsymbol{s_i}|\boldsymbol{V_i, c_i}) 
\end{equation}
where $q_s(\boldsymbol{s_i}|\boldsymbol{V_i},\boldsymbol{c_i})$ is defined after Eq.~(\ref{eq:2}).

For the coordinate refinement module, we aim to generate the precise coordinate of each point. Towards this goal, a MSE based loss is used as the distortion in the training procedure:
\begin{equation}
\setlength{\abovedisplayskip}{0.15cm}
\setlength{\belowdisplayskip}{0.1cm}
L_{mse} = \sum_{i} \left\|(x^{r}_i, y^{r}_i, z^{r}_i) - (x^{g}_i, y^{g}_i, z^{g}_i) \right\|_2^2
\end{equation}
where $(x^{g}_i, y^{g}_i, z^{g}_i)$ is the ground-truth coordinate of node $n_i$ before octree construction.

\begin{figure}[t]
\centering
\begin{minipage}[c]{0.9\linewidth}
% \centering
\flushright
\includegraphics[width=1\linewidth]{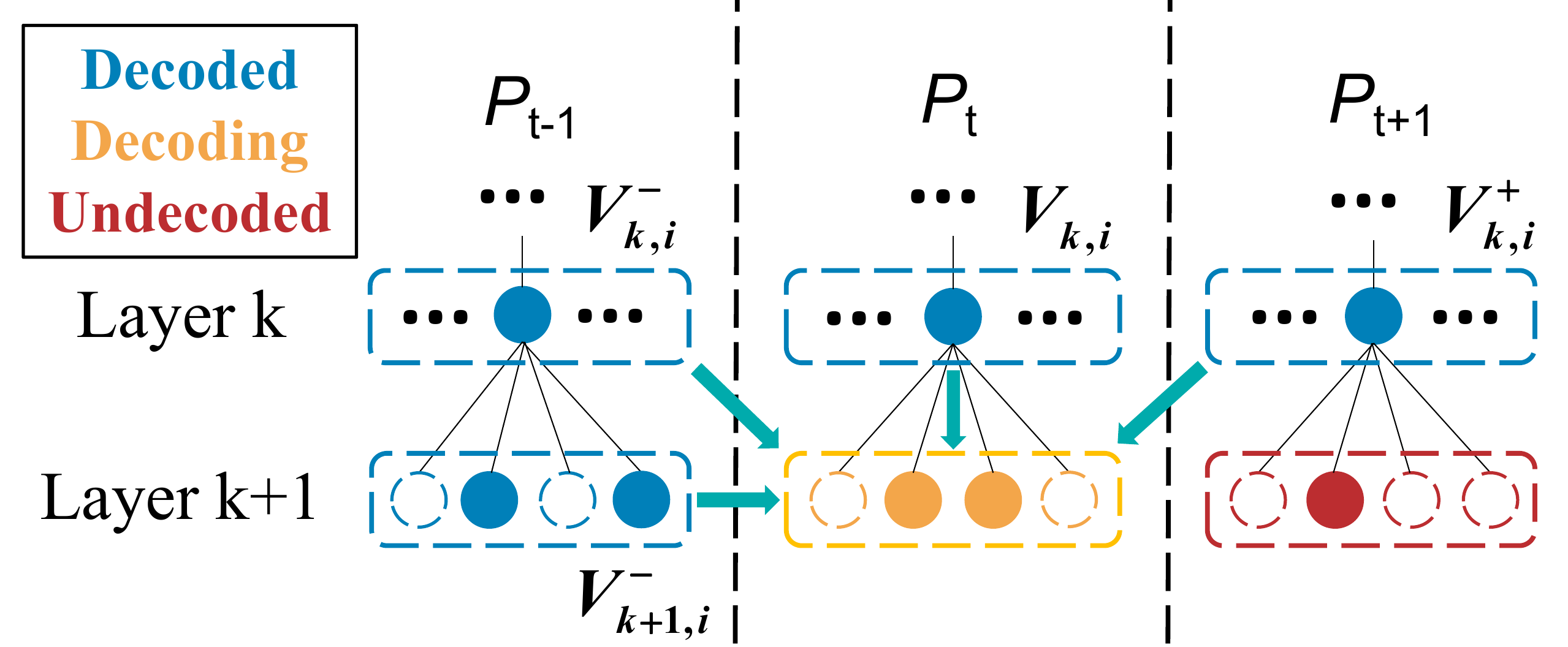}
\end{minipage}
\caption{Illustration of the local context representation for dynamic point clouds. Four voxel representations from the decoded layers(see the blue boxes) are used as local voxel context to estimate the probability distribution of each node in the current decoding layer(see the yellow box).}
\label{fig:Decoding}
\end{figure}

\subsection{Dynamic Point Cloud Compression}
\label{sec:dpcc}

The proposed point cloud compression framework is very general and can be readily extended for dynamic point cloud compression. 
Considering that each dynamic point cloud consists of a sequence of redundant point clouds that are captured at different time, it is necessary to exploit temporal information.

Specifically, given the point cloud $\mathcal{P}_t$ at time step $t$ and its neighbouring point clouds $\mathcal{P}_{t-1}$ and $\mathcal{P}_{t+1}$ at time step $t-1$ and $t+1$, we first align these point clouds to the same coordinate system based on their pose information of the sensor.
It should be mentioned that we decode \textit{all} point clouds from a sequence in a depth by depth fashion. 
Thus, when we decode the node $n_i$ at the $k$th depth level for the current point cloud $\mathcal{P}_t$, the voxel representation at the $k$th depth level from other two point clouds $\mathcal{P}_{t-1}$ and $\mathcal{P}_{t+1}$ is available. Meanwhile, the voxel representation at the $(k+1)$th depth level from the point cloud $\mathcal{P}_{t-1}$ is also available.

Here, as shown in Fig.~\ref{fig:Decoding}, we assume the local voxel representation for node $n_i$ at the $k$th depth level in $\mathcal{P}_t$ is $\boldsymbol{V_{k,i}}$ and the corresponding co-located local voxel representations at the $k$th depth level in $\mathcal{P}_{t-1}$ and $\mathcal{P}_{t+1}$ are $\boldsymbol{V^{-}_{k,i}}$ and $\boldsymbol{V^{+}_{k,i}}$, respectively.
Furthermore, the voxel representation at the $(k+1)$th depth level in the previous point cloud is denoted as $\boldsymbol{V^{-}_{k+1,i}}$.
To exploit temporal information in the entropy model, we adopt a simple fusion strategy and use the similar network architecture as shown in Figure~\ref{fig:dequantization} to extract the features from each local voxel representation.
Then we concatenate the features extracted from four different voxel representations (\textit{i.e.,} $\boldsymbol{V_{k,i}}$, $\boldsymbol{V^{-}_{k,i}}$, $\boldsymbol{V^{+}_{k,i}}$ and $\boldsymbol{V^{-}_{k+1,i}}$) and feed the aggregated feature to the \textit{Softmax} layer to produce the final probability distribution.
In our implementation, we set the size of $\boldsymbol{V_{k,i}}$, $\boldsymbol{V^{-}_{k,i}}$ and $\boldsymbol{V^{+}_{k,i}}$ as $9 \times 9 \times 9$, and set the size of $\boldsymbol{V^{-}_{k+1,i}}$ as $10 \times 10 \times 10$.
Finally, the coordinates from the reconstructed octree are refined by using the voxel context based method discussed in Section~\ref{sec:CR}.
It is noted that we only use the voxel representation $\boldsymbol{V_{k,i}}$ as context information for coordinate refinement as it is sufficient to use this voxel representation to achieve promising results.

\section{Experiments}

\begin{figure*}[t]
\centering
% 2 for ScanNet
\begin{minipage}[c]{0.3\textwidth}
\includegraphics[width=1\textwidth]{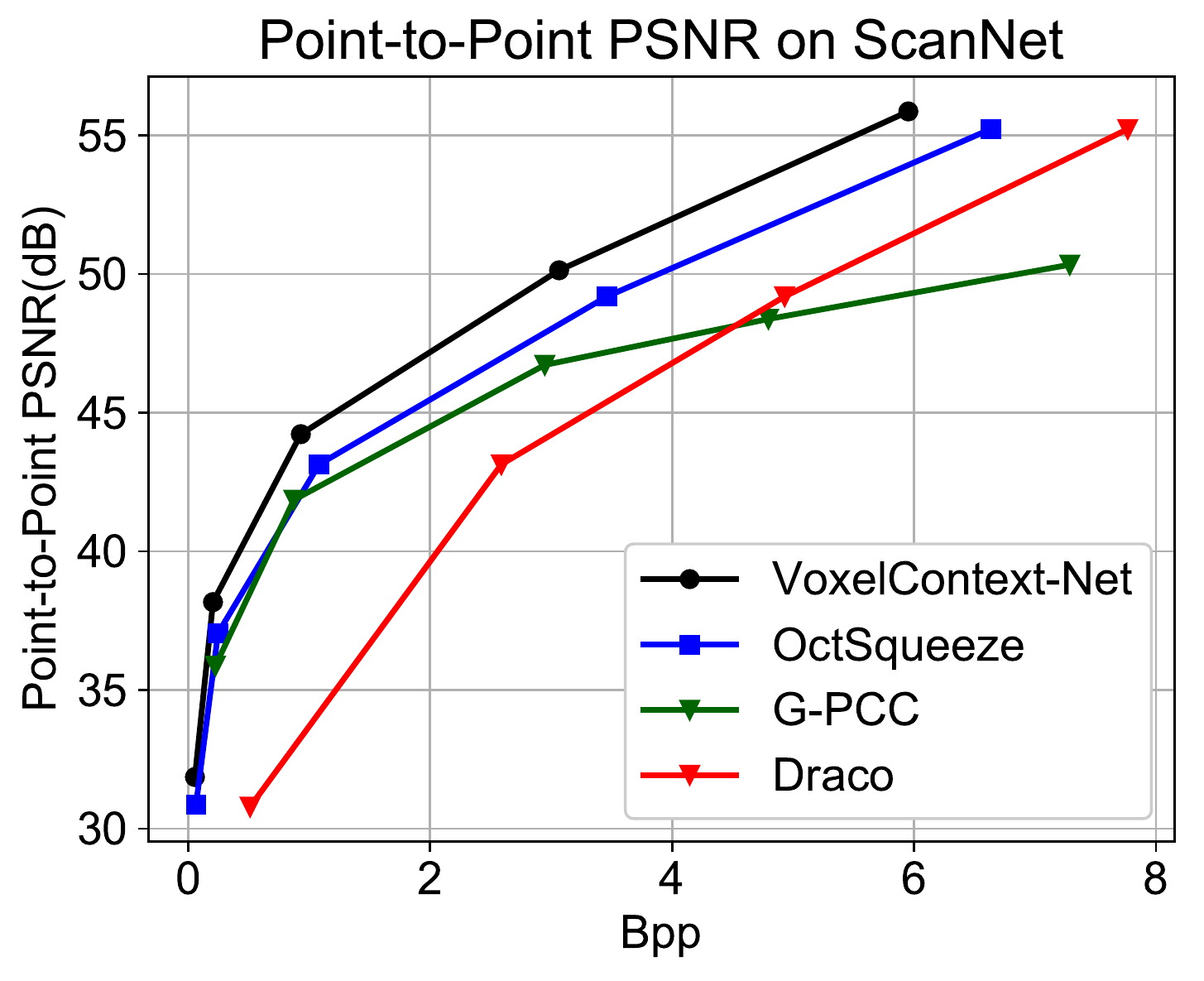}
\end{minipage}
\begin{minipage}[c]{0.3\textwidth}
\includegraphics[width=1\textwidth]{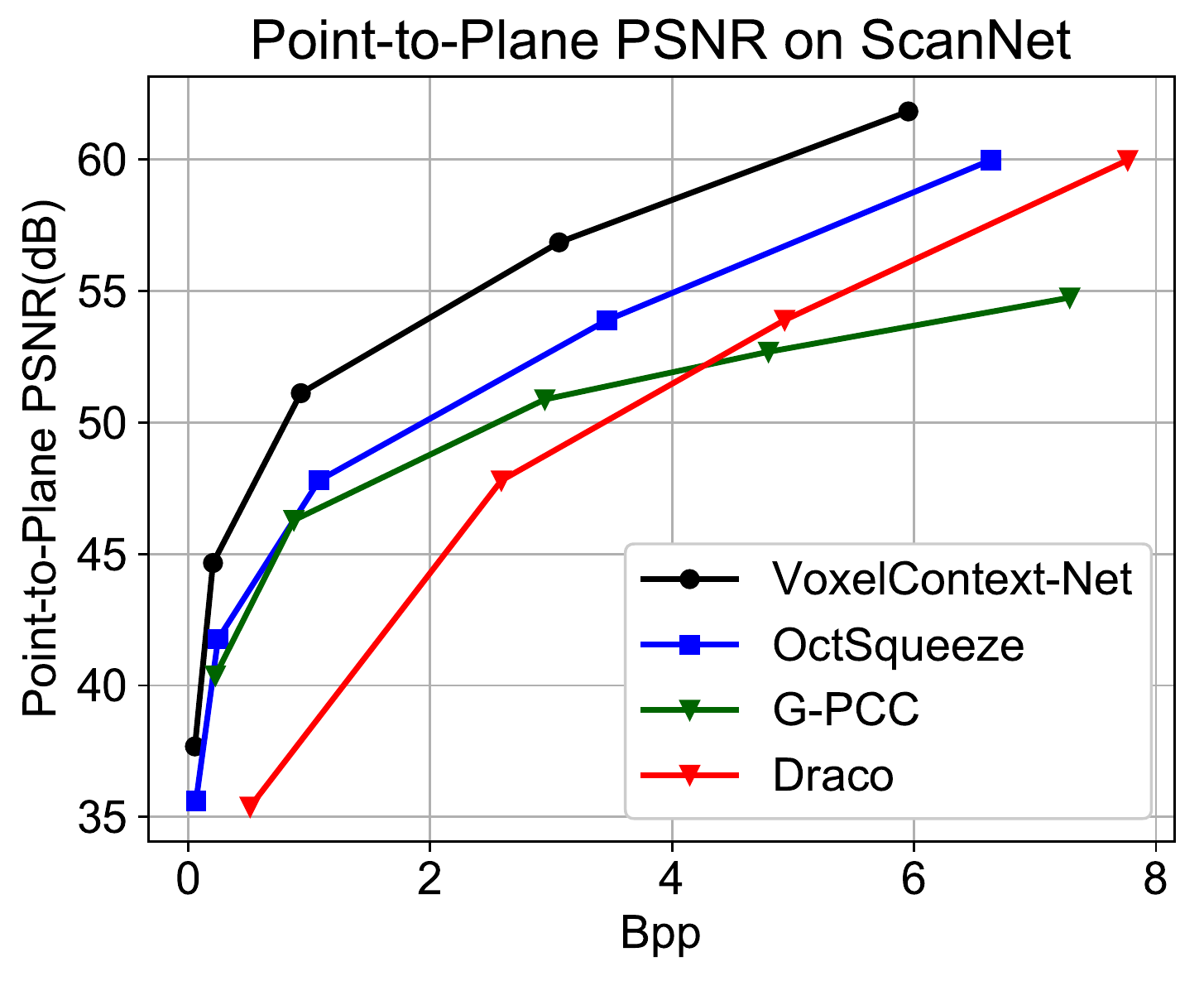}
\end{minipage}
\begin{minipage}[c]{0.3\textwidth}
\includegraphics[width=1\textwidth]{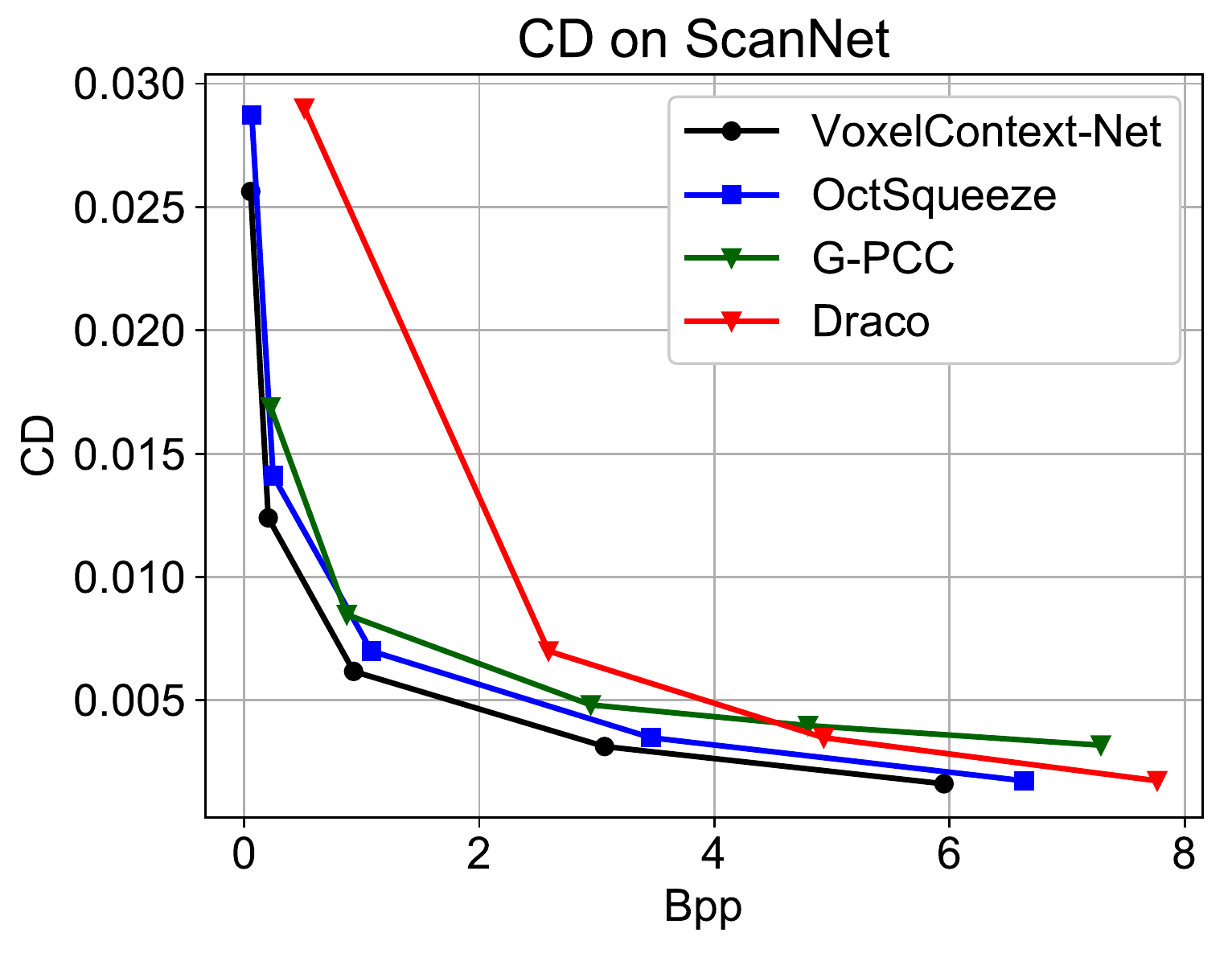}
\end{minipage}
% \vfill
\begin{minipage}[c]{0.3\textwidth}
\includegraphics[width=1\textwidth]{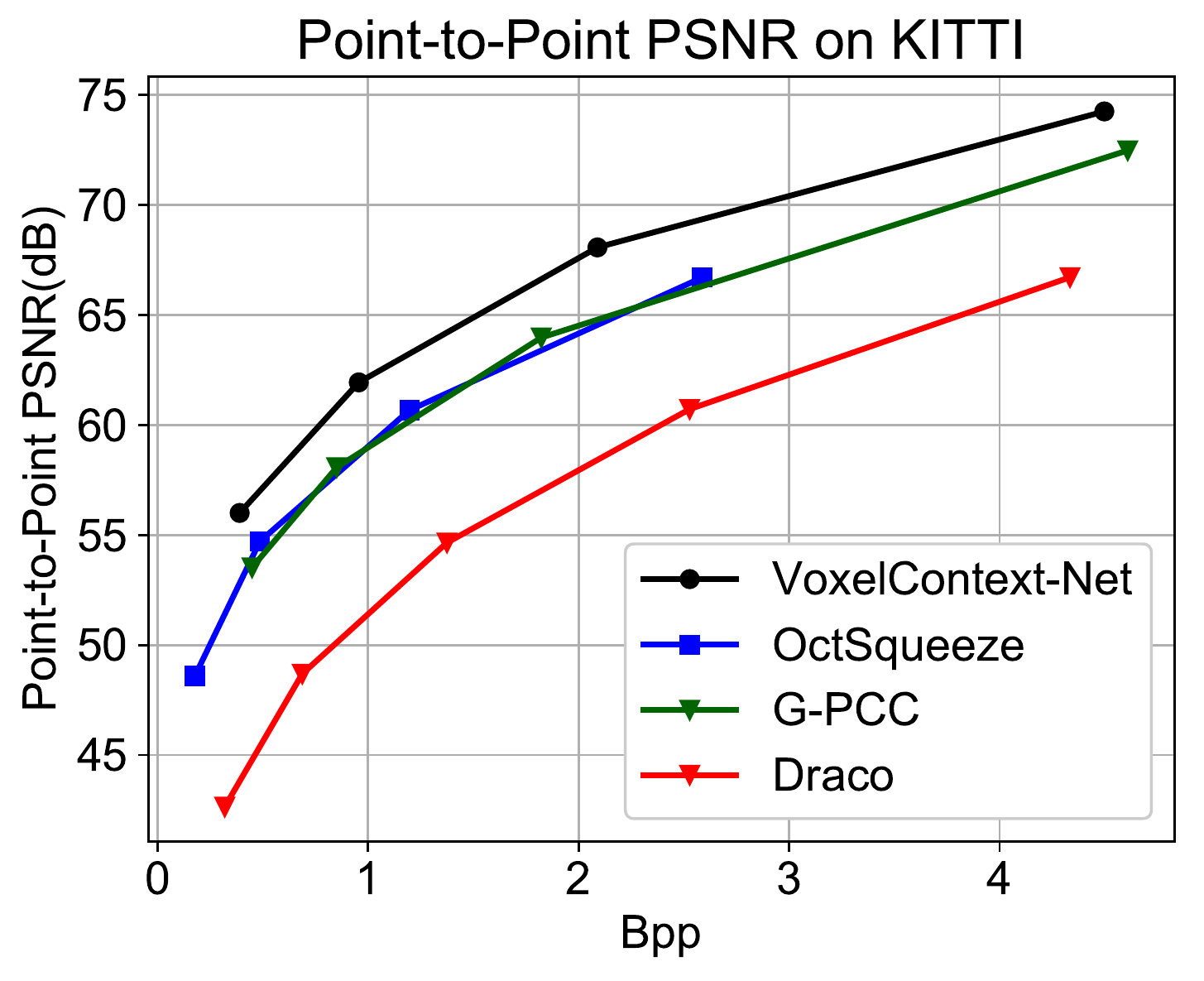}
\end{minipage}
\begin{minipage}[c]{0.3\textwidth}
\includegraphics[width=1\textwidth]{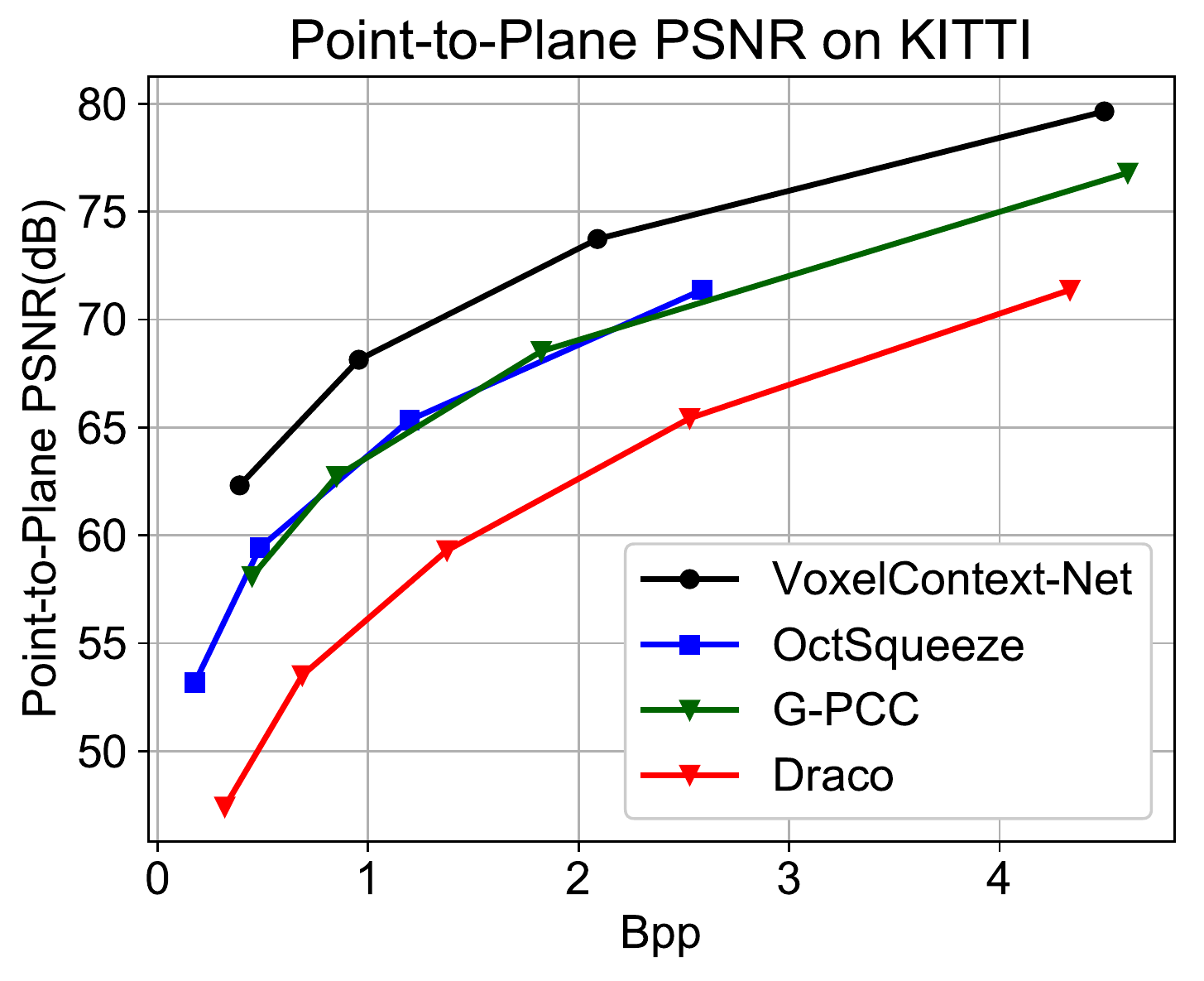}
\end{minipage}
\begin{minipage}[c]{0.3\textwidth}
\includegraphics[width=1\textwidth]{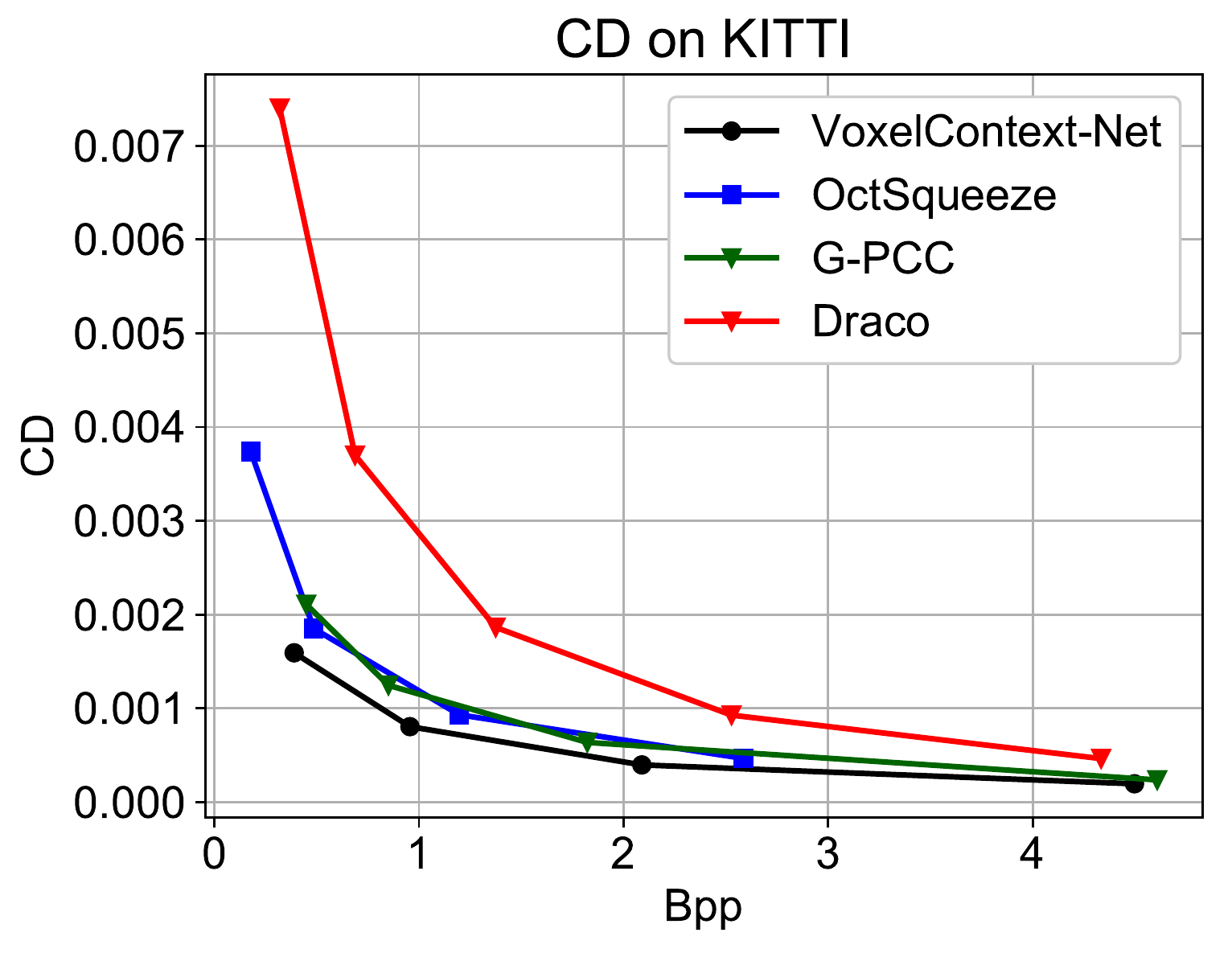}
\end{minipage}
% \caption{Results of different static point cloud compression methods on two benchmark datasets ScanNet \& Semantic KITTI.}
\caption{Results of different static point cloud compression methods on two benchmark datasets ScanNet \& Semantic KITTI.}
\label{fig:AllResults}
\end{figure*}

\begin{figure*}[t]
\centering
\begin{minipage}[c]{0.3\textwidth}
\includegraphics[width=1\textwidth]{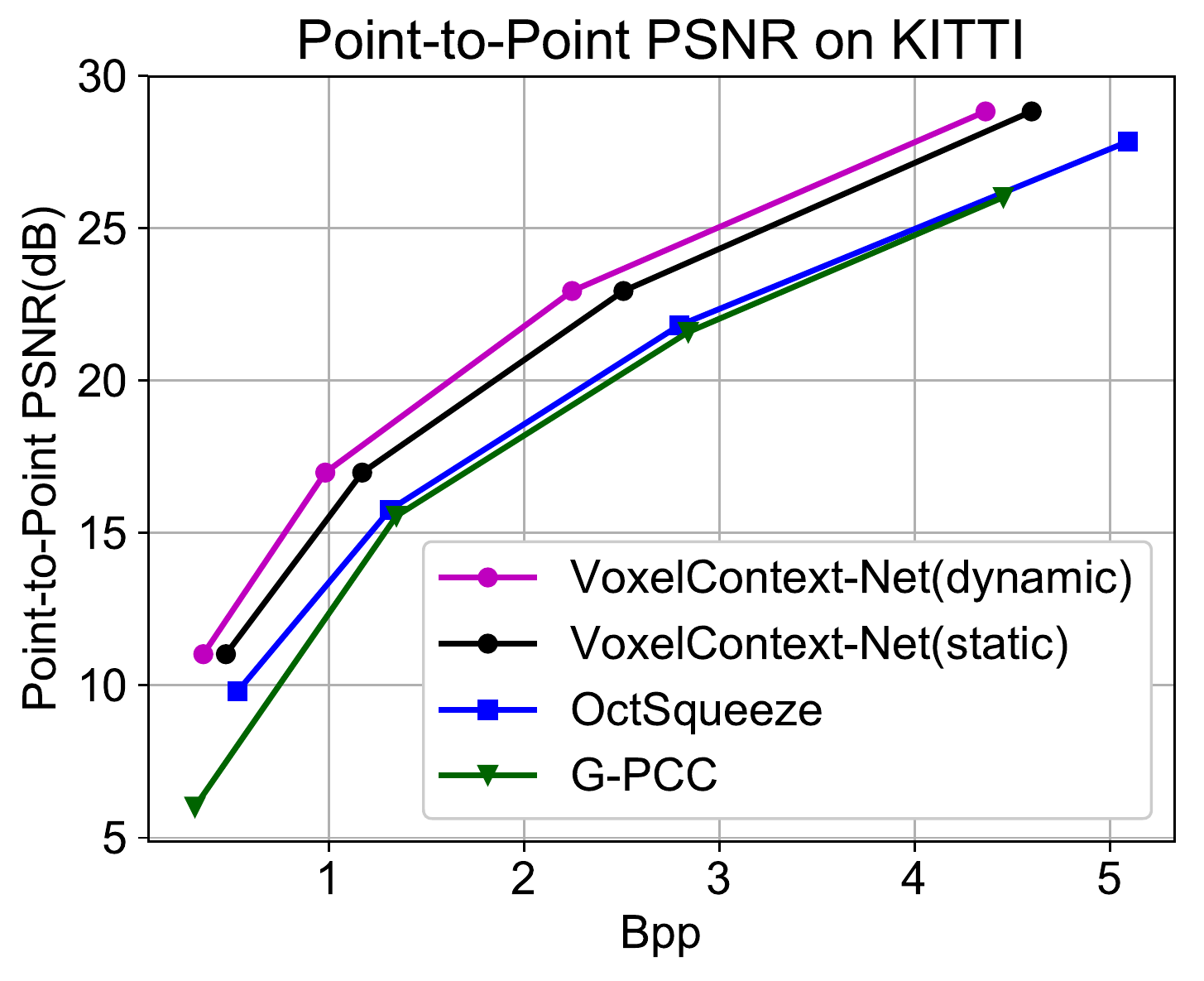}
\end{minipage}
\begin{minipage}[c]{0.3\textwidth}
\includegraphics[width=1\textwidth]{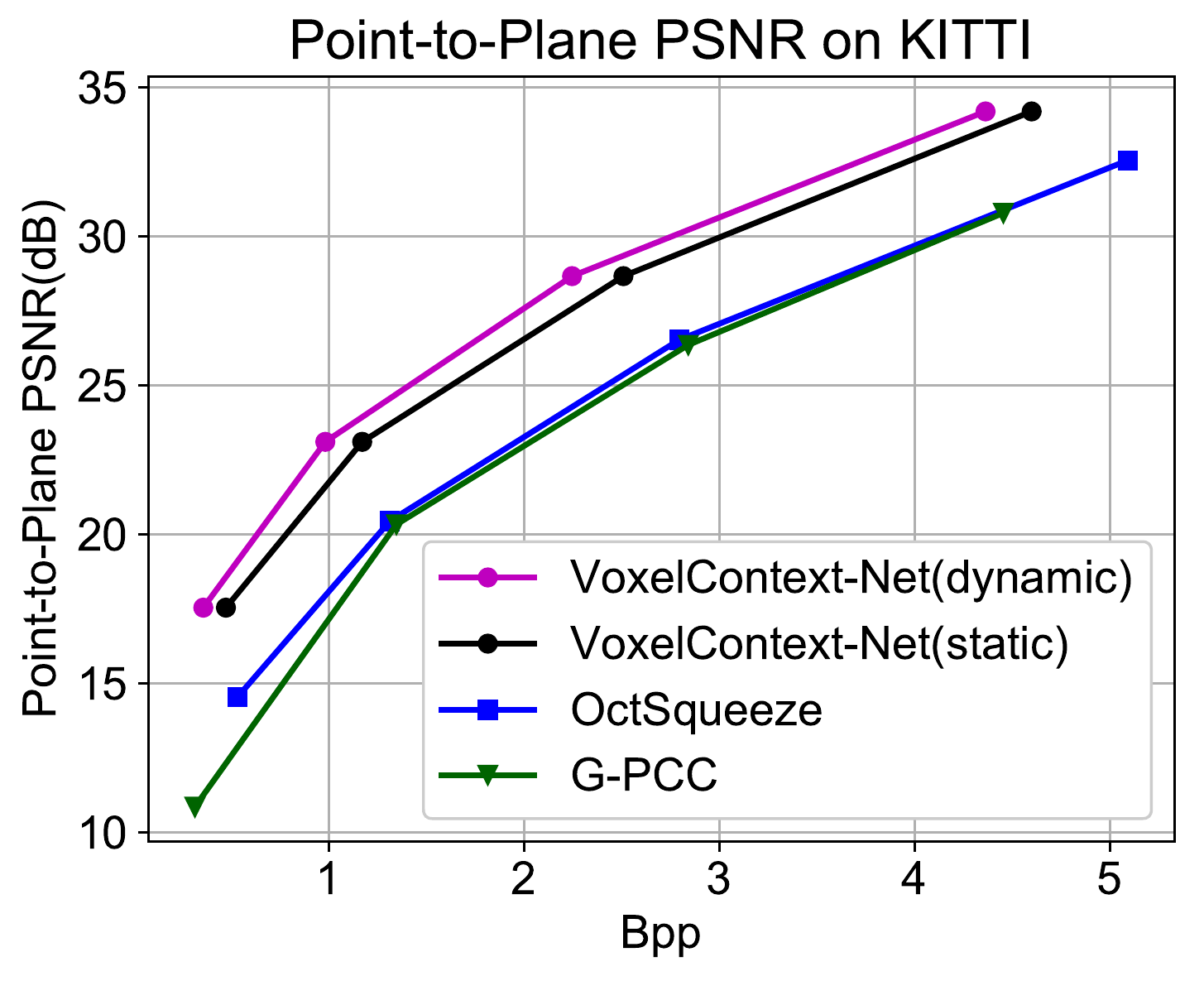}
\end{minipage}
\begin{minipage}[c]{0.3\textwidth}
\includegraphics[width=1\textwidth]{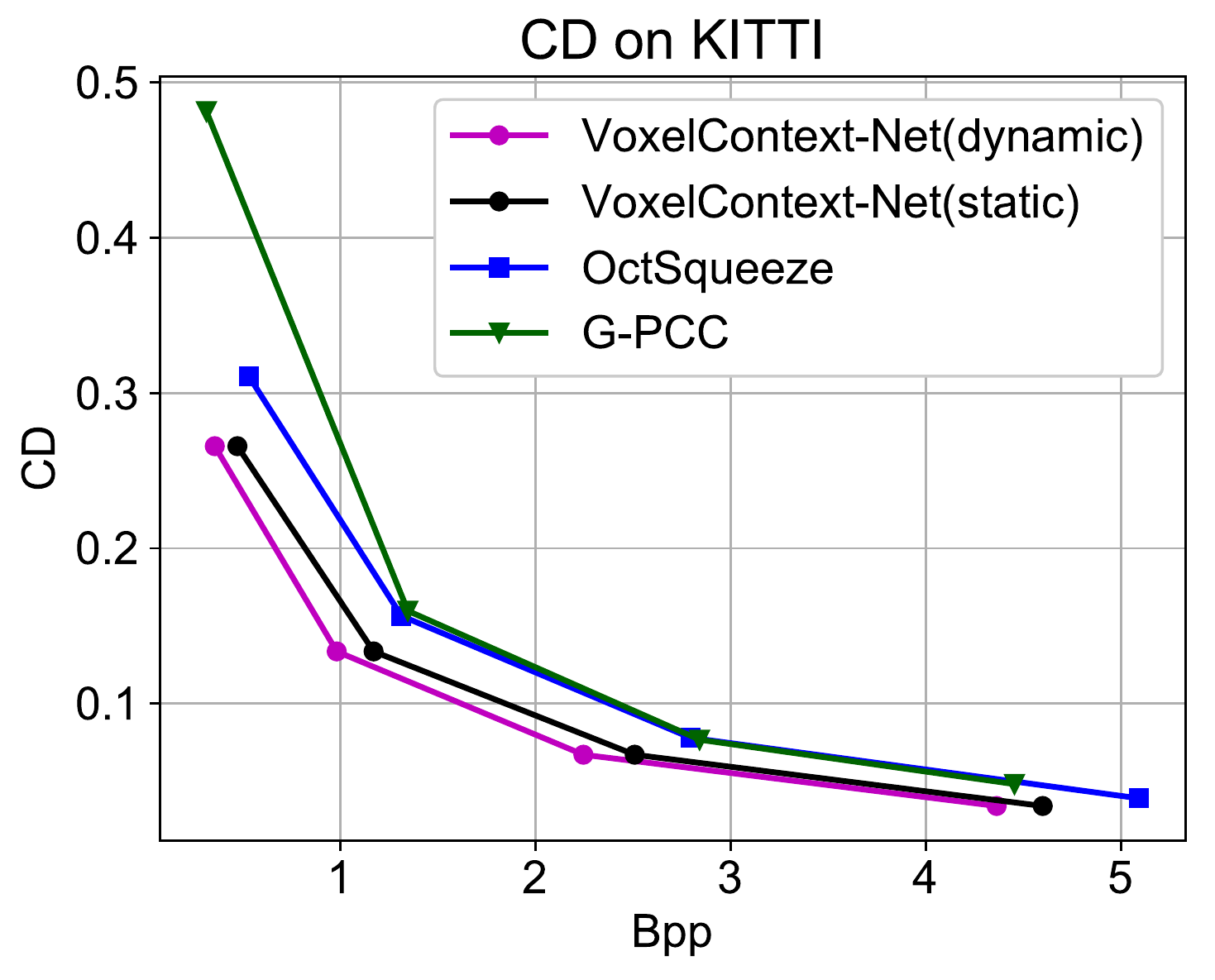}
\end{minipage}
\caption{Results of different methods for dynamic point cloud compression on the Semantic KITTI dataset.}
\label{fig:AllResults_dynamic}
\end{figure*}

\subsection{Datasets}

\noindent
\textbf{ScanNet}: ScanNet~\cite{dai2017scannet} is a large-scale dataset with a few dense point cloud sequences from the indoor scenario. It consists of 1,513 scans. In our experiment, 50,000 points are sampled from each scan.

\noindent
\textbf{Semantic KITTI}:  Semantic KITTI~\cite{geiger2012cvpr,behley2019iccv} is another large-scale dataset with several dense point cloud sequences captured from the self-driving scenario, which contains 23,311 scans with about 5 billion points. 
In our experiment, it is used for both static and dynamic point cloud compression.

For the ScanNet dataset,  we use the official training/testing split, which includes 1,201 point clouds for training and 312 point clouds for testing. For evaluating the static point cloud compression methods on the Semantic KITTI dataset, we follow the default setting, where 11 point cloud sequences are used for training and the other 11 sequences are used for testing.
For dynamic point cloud compression, considering that the ground-truth pose information is not available in the official testing sequences, in this work, we only use 11 training sequences for training and performance evaluation, in which 8 sequences are used for training while 3 sequences are used for performance evaluation.

\subsection{Experimental Details}

\noindent
\textbf{Baseline Methods.}
In our experiments, the two most representative hand-crafted point cloud compression methods, MPEG's standard point cloud compression method(`G-PCC')~\cite{schwarz2018emerging,gpcc} and Google's KD-tree based method(`Draco')~\cite{google} are used as the baseline algorithms. 
In addition, we also compare our method with the recent learning based static point cloud compression method OctSqueeze~\cite{Huang_2020_CVPR}. 
Since the source code of OctSqueeze is not publicly available, we re-implemented it by ourselves.

\noindent
\textbf{Training and Testing Strategy.}
In our training procedure for static point cloud compression, the maximum depth levels on ScanNet and SemanticKITTI are empirically set as 9 and 12, respectively. 
And the entropy model is optimized by using all nodes from the complete 9-level/12-level octree.
In the testing stage, we truncate the octree at different levels to evaluate our deep entropy models at different bitrates. 
It is noted that we train a coordinate refinement model at each depth level to reduce the distortion in the octree construction procedure.
For dynamic point cloud compression, we adopt the same training procedure.

We perform the experiments on the machine with one NVIDIA 2080TI GPU.
Our whole network is implemented based on PyTorch and it takes 3 days and 5 days to train the model, for the static and dynamic point cloud compression tasks, respectively.
In the training procedure, we use the Adam~\cite{kingma2014adam} optimizer and 
the learning rate is set as $1e-4$ for both entropy model and the coordinate refinement model.

\noindent
\textbf{Evaluation Metrics}
In our experiments, we use the Chamfer distance(CD)~\cite{fan2017point, huang20193d}, point-to-point PSNR and point-to-plane PSNR~\cite{tian2017geometric, mekuria2017performance}, where $p$ is set as 1 to measure the quality of the reconstructed point cloud.
We simply use bits per point(Bpp) as the compression ratio metric.
To compare different compression algorithms, we also include the BDBR~\cite{bjontegaard2001calculation} in our approach, which represents the average bitrate saving when the reconstructed quality (\textit{e.g.,} PSNR) of these methods are the same.

\begin{figure*}[!t]
\centering
% ScanNet
\begin{minipage}[c]{\textwidth}
\centering
\includegraphics[width=0.92\textwidth]{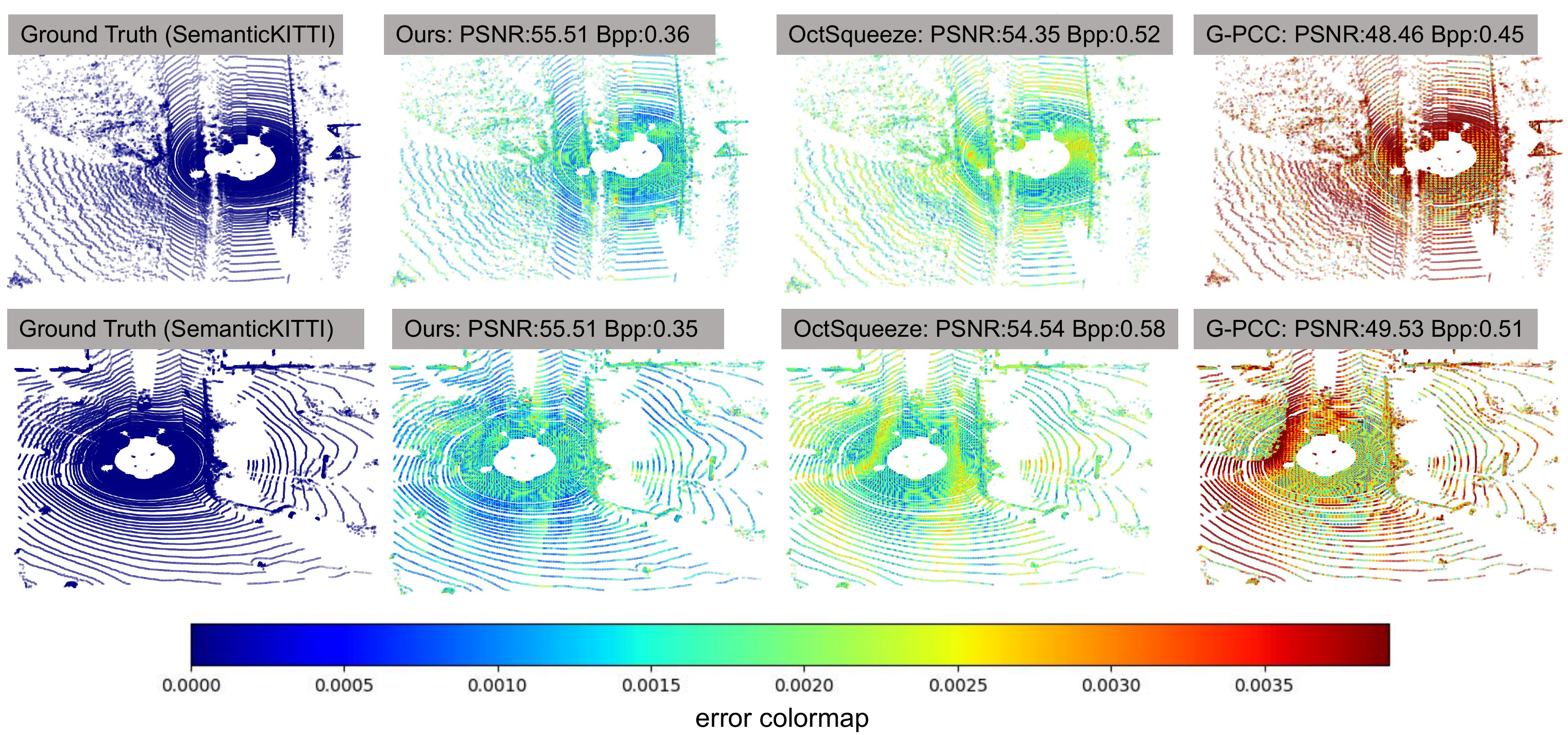}
\end{minipage}
\caption{Visualization of our VoxelContext-Net and other baseline methods for static point cloud compression on Semantic KITTI.}
\label{fig:Visualization}
\end{figure*}

\subsection{Experiment Results}

\noindent
\textbf{Results for Static Point Cloud Compression.}
The quantitative experimental results are provided in Table~\ref{tab:BDBR} and we use BDBR to evaluate the compression performance of different codecs, where G-PCC~\cite{schwarz2018emerging,gpcc} is used as the anchor algorithm.
It is observed that our approach~(\textit{i.e.,} VoxelContext-Net) saves 43.66\% bitrate on the ScanNet dataset when compared with G-PCC, while the corresponding bitrate saving for OctSqueeze is only 15.00\%. 
Similar results are also observed on the Semantic KITTI dataset, where our approach achieves 31.15\% bitrate saving, while OctSqueeze only saves 2.13\% bitrates.
These experimental results clearly demonstrate our approach outperforms the state-of-the-art learning based compression algorithm and the traditional codecs like G-PCC.

For static point cloud compression, the corresponding rate-distortion curves are provided in Figure~\ref{fig:AllResults}. Our approach achieves better compression performance, especially at high bitrates.
For example, our approach has more than 2dB improvement over OctSqueeze on the ScanNet dataset when the bpp is 4.

In Figure~\ref{fig:Visualization}, we take the Semantic Kitti dataset as an example to provide the qualitative results. 
We observe that the errors between the ground-truth point clouds and our reconstructed point clouds are also smaller when compared with the baseline methods OctSqueeze and G-PCC.
For example, the bpp of our approach is 0.36 and the corresponding PSNR is 55.51dB, while OctSqueeze achieves a lower PSNR(54.35dB) when using more bits(0.52bpp).

\begin{figure}[t]
\centering
% ScanNet
\begin{minipage}[c]{0.30\textwidth}
\includegraphics[width=1\textwidth]{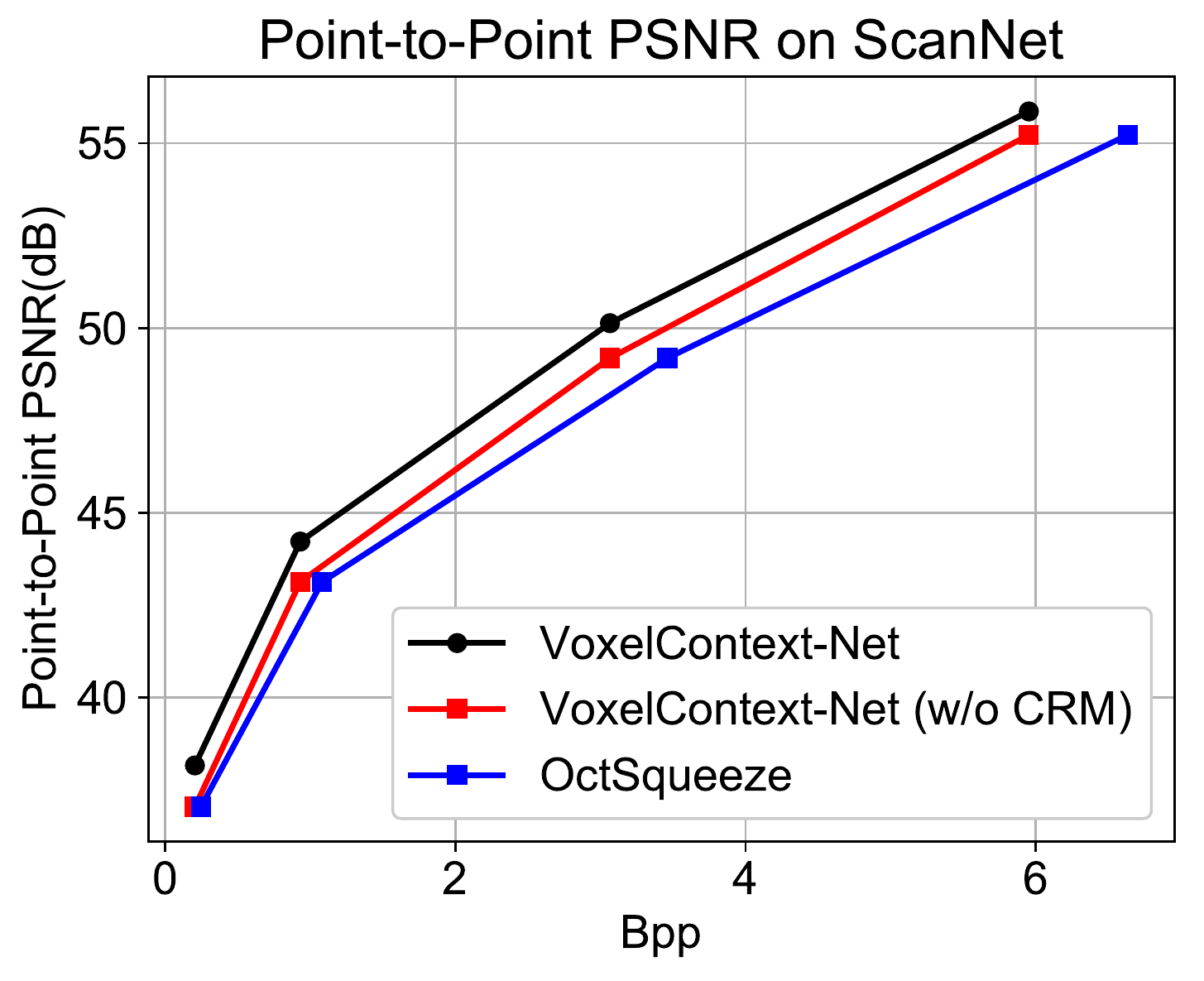}
\end{minipage}
\caption{Ablation study on the Scannet dataset.}
\label{fig:AblationContext}
\end{figure}

\noindent
\textbf{Results for Dynamic Point Cloud Compression.}
For dynamic point cloud compression, we provide the quantitative results in Table~\ref{tab:BDBR2} and Figure~\ref{fig:AllResults_dynamic}. It is observed that our dynamic point cloud compression method (\textit{i.e.,} Ours(dynamic)) outperforms the baseline methods G-PCC and OctSqueeze by a large margin. For example, Ours(dynamic) saves 38.10\% bitrates when compared with the anchor algorithm G-PCC while the corresponding saving is only 5.01\% for the OctSqueeze method. Furthermore, when compared with Ours(static), Ours(dynamic) saves additional 11.11\% bitrates (-26.99\% vs. -38.10\%) on Semantic KITTI. 
Considering that the only difference between Ours(static) and Ours(dynamic) is the additional temporal voxel context in the entropy model~(see Section~\ref{sec:dpcc} and Figure~\ref{fig:Decoding}), the results demonstrate it is effective to exploit the temporal information for dynamic point cloud compression.

Since our approach requires the pose information of the sensor when performing point cloud alignment, we also provide the results on the KITTI dataset when our approach uses the estimated pose information~\cite{yin2020caelo}, instead of the ground truth pose information. 
The experimental result in Table~\ref{tab:BDBR2} demonstrates that our approach with the estimated pose information (\textit{i.e.,} Ours*(dynamic)), also saves 41.77\% bitrate and still outperforms the existing baseline methods. 
It is noted that Ours*(dynamic) performs even better than Ours(dynamic). One possible explanation is that the ground-truth pose represents motion information of the sensors while the estimated pose information is calculated based on the motion between actual point clouds, which may be more useful for the compression task.

\subsection{Ablation Study and Analysis}

In this section, we take our VoxelContext-Net for static point cloud compression on the ScanNet dataset as an example to perform the ablation study.

\noindent
\textbf{Effectiveness of Our Proposed Components.}
To demonstrate the effectiveness of our proposed two components, we consider a simplified version of our approach by removing the coordinate refinement module, which is referred to as VoxelContext-Net (w/o CRM). 

Based on the experimental results~(see the purple curve) in Figure~\ref{fig:AblationContext}, we have the following two observations.
First, when compared with the octree based entropy model in OctSqueeze~\cite{Huang_2020_CVPR}, our approach using the local voxel context boosts the performance and saves an additional 14.81\% bitrate in terms of BDBR.
It shows it is more effective to use context information provided by the local voxel representation rather than that extracted from the parent and child nodes~\cite{Huang_2020_CVPR}.
Second, when comparing our full model~(see the black curve) and the simplified model~(see the purple curve), it is observed that the coordinate refinement procedure further improves the performance and saves 21.50\% bitrates in terms of BDBR. 
Therefore, it is beneficial to reduce the distortion from the octree construction procedure by using our proposed coordinate refinement module.

\begin{table}[!t]
    \begin{center}
            \caption{BDBR(\%) results of our method Ours(static) and two baseline algorithms Draco~\cite{google} and OctSqueeze~\cite{Huang_2020_CVPR} when compared with G-PCC~\cite{schwarz2018emerging,gpcc} on two benchmark datasets for static point cloud compression.}
% \resizebox{\textwidth}{4mm}{
\resizebox{0.77\columnwidth}{!}{
\setlength{\tabcolsep}{1mm}{
    \begin{tabular}{|c|c|c|c|c|}
    \hline
        Methods & Draco & OctSqueeze &  Ours(static)\\
    \hline
        ScanNet & +133.32 & -15.00 & -43.66 \\
    \hline
        Semantic KITTI & +138.58 & -2.13 &-31.15\\
    \hline
    \end{tabular}}}
    \label{tab:BDBR}
    \end{center}
\end{table}

\begin{table}[!t]
    \begin{center}
            \caption{BDBR(\%) results of our methods and the baseline algorithm OctSqueeze~\cite{Huang_2020_CVPR} when compared with G-PCC on Semantic KITTI for dynamic point cloud compression. In Ours(dynamic) and Ours*(dynamic), we use the ground-truth and the estimated pose information, respectively.}
\resizebox{0.9\columnwidth}{!}{
\setlength{\tabcolsep}{1mm}{
    \begin{tabular}{|c|c|c|c|c|}
    \hline
        OctSqueeze &  Ours(static) & Ours(dynamic) & Ours*(dynamic) \\
    \hline
        -5.01 & -26.99 & -38.10 & -41.77 \\
    \hline
    \end{tabular}}}
    \label{tab:BDBR2}
    \end{center}
\end{table}

\noindent
\textbf{Voxel Size.}
In our implementation, the size of local voxel representation for each node is empirically set as $9 \times 9 \times 9$. As shown in Table~\ref{tab:bpp}, we provide more experimental results when the size varies. It is noted that the performance can be boosted by increasing the resolution of the local voxel representations. For example, when compared with our entropy model with the voxel size as $5 \times 5 \times 5$, our proposed approach saves 2.5\% bitrates when the voxel size becomes $9 \times 9 \times 9$ at the depth level 9. Considering that the computational complexity will also increase by using larger voxel sizes, we choose $9 \times 9 \times 9$ as the default size in our approach for better trade-off between compression performance and computational complexity.

\noindent
\textbf{Computational Complexity} 
In Table~\ref{tab:Runtime}, we provide the decoding time of different methods at various bitrates on two datasets. Our method is faster than G-PCC~\cite{schwarz2018emerging, gpcc}. While it is slower than OctSqueeze~\cite{Huang_2020_CVPR}, we achieve better compression performance~(see Fig.~\ref{fig:AllResults} and Table~\ref{tab:BDBR}). The total number of parameters in our approach is 2.15M.

\noindent
\textbf{Experimental Results for the Downstream Tasks}
We evaluate the impact of different point cloud compression methods for two downstream tasks (\textit{i.e.,} object detection and semantic segmentation).
Specifically, we use VoteNet~\cite{qi2019deep} as the object detection method and PointNet++~\cite{qi2017pointnet++} as the semantic segmentation method. We train these models based on the uncompressed point clouds from the official training dataset of ScanNet.
In the evaluation stage, the reconstructed point clouds at different bitrates are fed into the detection or the segmentation method. 
For the object detection task, we employ mAP@0.25 to measure the detection accuracy.
Following the setting in~\cite{Huang_2020_CVPR}, for the semantic segmentation task, we adopt the intersection-over-union (IOU) score for performance evaluation, which is computed based on the ground truth labels for each voxel.

\begin{table}[!t]
    \begin{center}
            \caption{Results of our deep entropy model when using various sizes of local voxel representations on the ScanNet dataset.}
            \resizebox{0.9\columnwidth}{!}{%            
    \begin{tabular}{|c|c|c|c|c|}
    \hline
        \multirow{2}{*}{Size} & \multicolumn{4}{c|}{Bpp}\\
    % \hline 
    \cline{2-5}
        & Depth=6  &Depth=7 & Depth=8 & Depth=9\\
    \hline
        $5 \times 5 \times 5$  & 0.207 & 0.950 & 3.137 & 6.119\\
    \hline
        $7 \times 7 \times 7$  & 0.205 & 0.937 & 3.090 &6.009 \\
    \hline
        $9 \times 9 \times 9$  &0.204 & 0.930 & 3.065& 5.952\\
    \hline
        $11 \times 11 \times 11$ & 0.205 & 0.935 & 3.061 & 5.932\\
    \hline
    \end{tabular}
}
    \label{tab:bpp}
    \end{center}
\end{table}

\renewcommand \linespread{0.4}
\renewcommand\cellset{\renewcommand\arraystretch{0.85}
\setlength\extrarowheight{0pt}}
\begin{table}[t]
\setlength{\abovecaptionskip}{-0.5cm}
\setlength{\belowcaptionskip}{-1.5cm}
\caption{Decoding time(ms) on ScanNet/KITTI at five/four different bitrates reported in Figure~\ref{fig:AllResults}~(from low to high).}
\resizebox{0.96\columnwidth}{!}{
\setlength{\tabcolsep}{1mm}{
    \begin{tabular}{|c|c|c|c|c|}
    \hline
        Methods & Ours & OctSqueeze &  G-PCC\\
    \hline
        \makecell[c]{ScanNet} & 49/52/64/80/109 & 6/6/7/7/7 & 306/309/316/324/348 \\
    \hline
        \makecell{KITTI} & 52/58/78/90 & 6/7/7/8 & 578/649/734/768 \\
    \hline
    \end{tabular}
}
}
    \label{tab:Runtime}
\end{table}

\begin{figure}[!t]
\centering
% 2 for ScanNet
\begin{minipage}[c]{0.48\linewidth}
\includegraphics[width=1\textwidth]{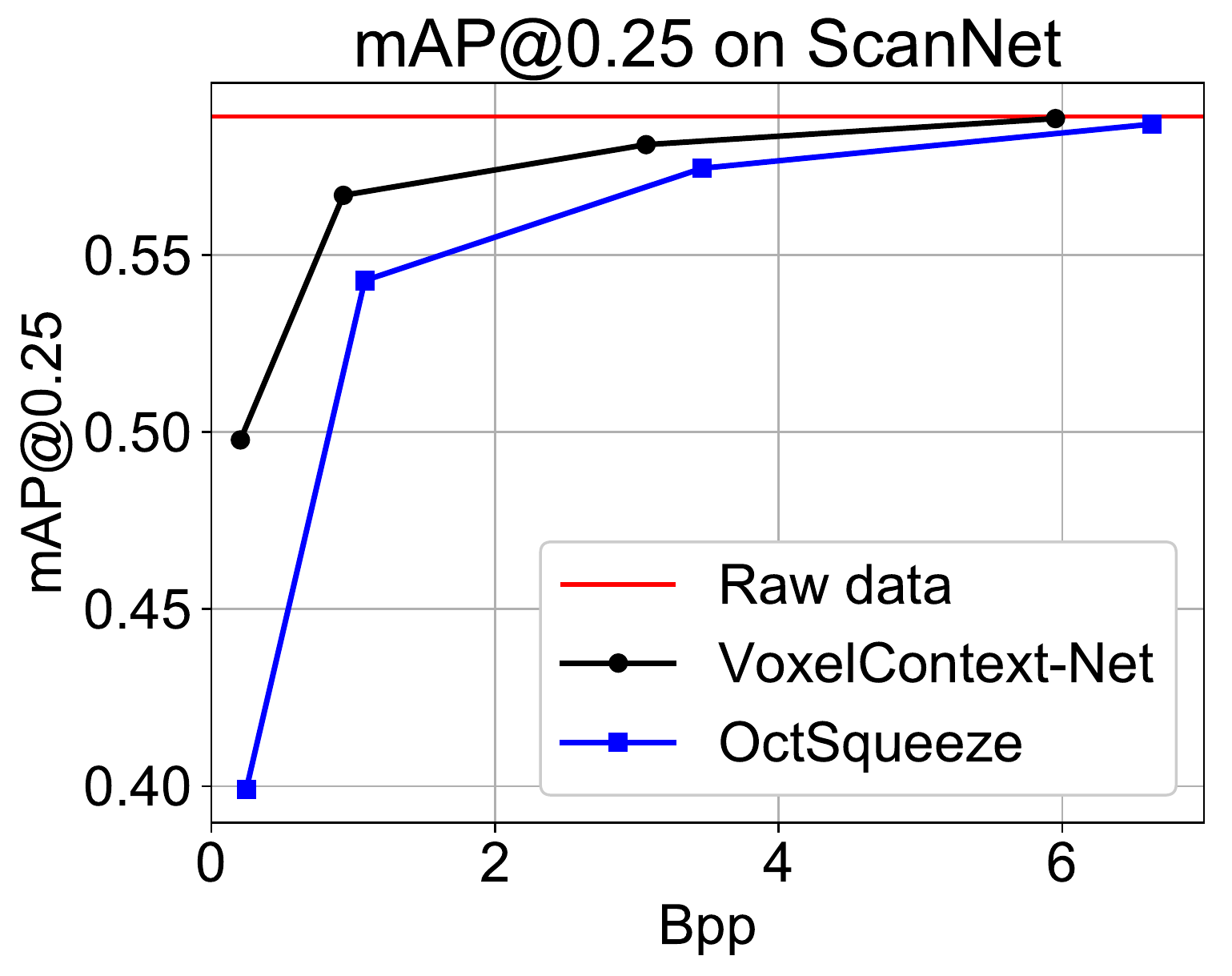}
\end{minipage}
\begin{minipage}[c]{0.48\linewidth}
\includegraphics[width=1\textwidth]{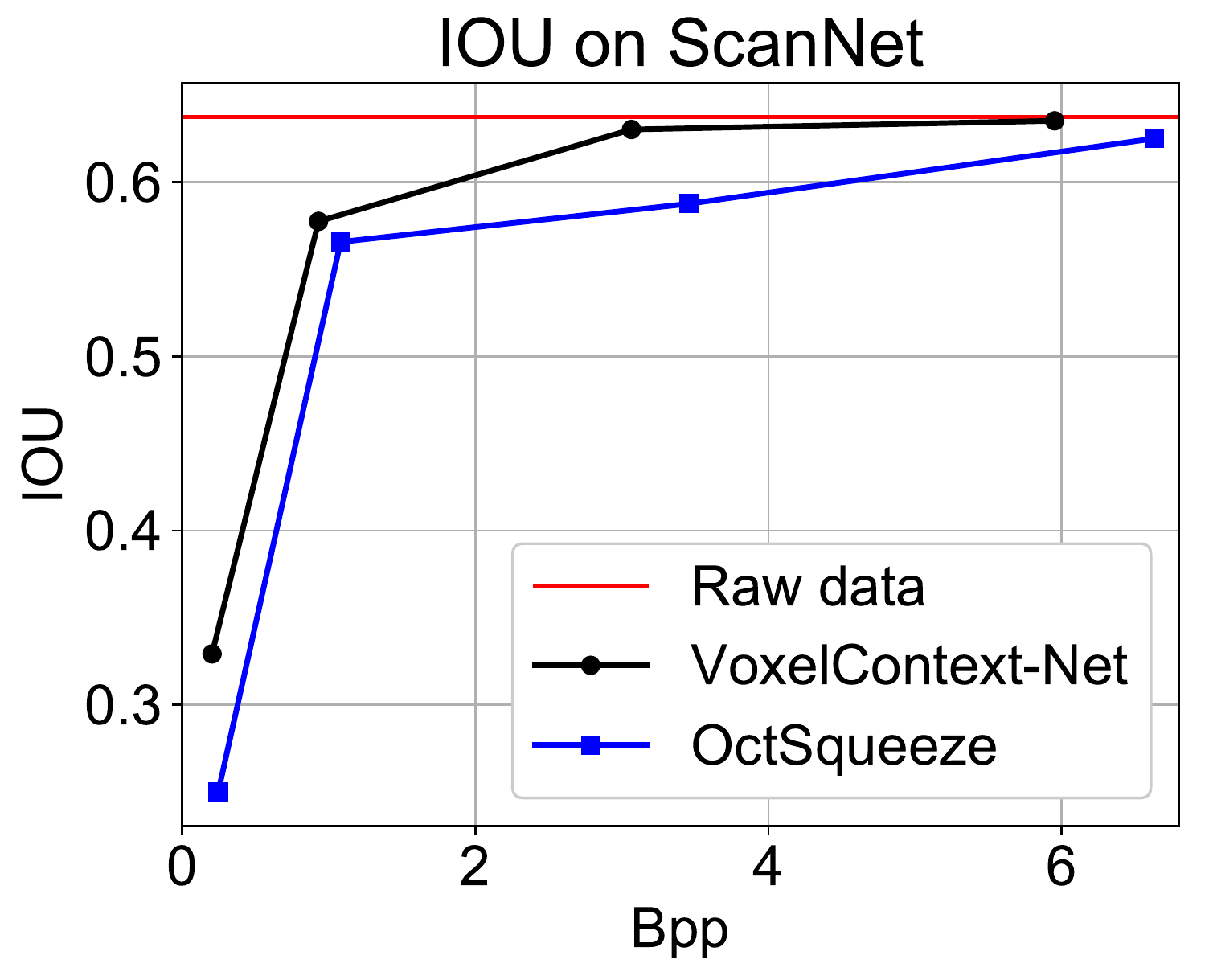}
\end{minipage}
\caption{The results for the two downstream tasks~(\textit{i.e.,} object detection~(left) and semantic segmentation~(right)) on ScanNet.}
\label{fig:Downstream}
\end{figure}

The experimental results are shown in Fig~\ref{fig:Downstream}.
Specifically, we provide the compression ratios~(i.e., bpps) and the corresponding detection/segmentation accuracy~(i.e., mAP/IOU) when using the decoded point clouds from our VoxelContext-Net and the baseline method OctSqueeze as the input to the detection/segmentation method.
At any given bpp, it is obvious that the detection/segmentation results based on the decoded point clouds from our VoxelContext-Net are higher, especially at low-bitrates.
It demonstrates that the reconstructed point clouds from our VoxelContext-Net are more useful for the downstream tasks, like object detection or semantic segmentation.

\section{Conclusion}
In this work, we have proposed a new learning based point cloud geometry compression framework by exploiting the local voxel representation for each node in the octree structured point cloud.
Specifically, we propose a new deep entropy model to losslessly compress the symbols of octree nodes and a new coordinate refinement module for reconstructing high-quality point clouds at the decoder side. Our simple and effective approach is applied to both static and dynamic point cloud compression and our method has achieved the state-of-the-art compression performance on two benchmark datasets.

\vspace{+1mm}

\noindent\textbf{Acknowledgement}
This work was supported by the National Key Research and Development Project of China (No. 2018AAA0101900). 

{\small

\bibliographystyle{ieee_fullname}
%\bibliography{mybib}
}

\end{document}